\documentclass[preprint,12pt]{elsarticle}

\usepackage{graphicx}
\graphicspath{{./figures/}}
\usepackage{amssymb}
\usepackage{amsmath}
\usepackage{url}
\usepackage{hyperref}
\usepackage{tabularx}
\usepackage{subfig}
\usepackage{booktabs} 
\usepackage[numbers]{natbib}

\usepackage{lineno}

\newcommand{\etal}{\textit{et al.}}

\begin{document}
\begin{frontmatter}

\title{ESAD: Endoscopic Surgeon Action Detection Dataset}



\author[label1]{Vivek Singh Bawa\fnref{label3}}
\ead{vsingh@brookes.ac.uk}
\author[label2]{Gurkirt Singh\fnref{label3}}
\author[label1]{Francis Kaping'A}
\author[label1]{Inna Skarga-Bandurova}
\author[label4]{Alice Leporini}
\author[label4]{Carmela Landolfo}
\author[label4]{Armando Stabile}
\author[label5]{Francesco Setti}
\author[label5]{Riccardo Muradore}
\author[label4]{Elettra Oleari}
\author[label1]{Fabio Cuzzolin}

\address[label5]{University of Verona, Italy}
\address[label4]{San Raffaele Hospital, Milan, Italy}
\address[label1]{Oxford Brookes University, Oxford, UK\fnref{label7}}
\address[label2]{Computer Vision Lab, ETH Zurich, Switzerland}

\fntext[label3]{Authors have equal contribution}

\begin{abstract}
In this work, we take aim towards increasing the effectiveness of surgical assistant robots.
We intended to make assistant robots safer by making them aware about the actions of surgeon, so it can take appropriate assisting actions. 
In other words, we aim to solve the problem of surgeon action detection in endoscopic videos. 
To this, we introduce a challenging dataset for  surgeon action detection in real world endoscopic videos. 
Action classes are picked based on the feedback of surgeons and annotated by medical professional. 
Given a video frame, we draw bounding box around surgical tool which is performing action and label it with action label.
Finally, we present a frame-level action detection baseline model based on recent advances in object detection.
Results on our new dataset show that our presented dataset provides enough interesting challenges for future method and it can serve as strong benchmark corresponding research in surgeon action detection in endoscopic videos.
\end{abstract}


\end{frontmatter}


\section{Introduction}
\label{S:1}
Minimally Invasive Surgery (MIS) is a very sensitive medical procedure. A general MIS surgical procedure involves two surgeons: main surgeon and assistant surgeon. Success of a MIS procedure depends upon multiple factors, such as, attentiveness of main surgeon and assistant surgeon, competence of surgeons, effective coordination between the main surgeon and assistant surgeon etc. 

According to Lancet Commission, each year 4.2 million people die within 30 days of surgery \cite{nepogodiev2019global}. Another study at John Hopkins University states that 10\% of total deaths in USA are due to medical error \cite{jhustudy}. There is no definite measure to compute and predict the risk factor involving surgeons. This make it very critical to monitor the set of action performed by surgeons in real time, so that, any  unfortunate event can be avoided.

Artificial Intelligence is being used in a lot of applications where human error has to be mitigated. The proposed dataset is another step in same direction. To make the surgical procedure safer, we should be able to identify and track the actions of main as well as assistant surgeon. This dataset is developed with the assistance of medical professionals as well as expert surgeon. More details of the data set can be found in section \ref{sec:dataset}.

Although there a lot of datasets for different action detection task computer vision. But there is no existing dataset for action detection in medical computer vision, specifically for MIS surgeries. Given the complexity of the scene and difficulty in the detection of surgeon action, this dataset will pave a path forward and set a benchmark for the medical computer vision research community. The task of action detection in medical scenario is a lot different from the general computer vision (more discussion in section \ref{sec:problem}), hence all the standard computer vision algorithm can not be directly deployed. The dataset will also lay the foundation for more robust algorithms which will be used in future surgical systems to accomplish tasks, such as, autonomous assistant surgeons, surgeon feedback systems, surgical anomaly detection, and so on.

\section{Literature review} \label{sec:literature}
Action detection or activity analysis for medical images is an under explored field. Hence most of the literature in this field will be borrowed from general activity analysis literature. The earlier works like \cite{petlenkov2008application} use hand motion of the surgeon to recognize the action preformed. Voros~\etal~\cite{voros2008towards} uses motion of tools to detect the point of interaction between tool and the organ. Kocev~\etal~\cite{kocev2014projector} uses point could generated using Microsoft Kinect camera to build the augmented reality model of real time actions performed by surgeon. In~\cite{van2019weakly}, authors used weakly supervised approach based on Gaussian Mixture Models (GMM) to recognize the surgeon actions. The approach was not developed for real surgical images and only recognised actions with assumption of one action per frame. Azari~\etal~\cite{azari2019using} use video of surgeon hand motion to prediction the surgical maneuvers. Li~\etal~\cite{li2016subaction} uses sub-action categories for early stage prediction of main surgical actions.

Most of the literature we will be borrowing from the human activity detection problem. In general, there are two types of activity analysis methods: static and dynamic. Static methods only have spatial information (image data) without any temporal context to current frame \cite{singh2017online, gkioxari2015contextual, peng2016multi}. The dynamic activity detection methods use video data which given temporal context to the motion or structure under the observation \cite{saha2017amtnet, kalogeiton2017action, hou2017end, hou2017tube}. 

Singh~\etal~\cite{singh2017online} used Single Shot multi-box Detector (SSD) \cite{liu2016ssd} to detect the activity in the frame. SSD is a very successful algorithm in object detection which predicts the object bounding boxes in a single shot, making it one of the fastest detection algorithms available. \cite{gkioxari2015contextual} used RCNN for the region proposal and the these proposals are used to learn the context information to produce a more accurate activity class. Saha~\etal~\cite{saha2017amtnet} proposed activity detection module called as 3D-RPN (3 dimensional region proposal network) which uses spatial as well as temporal information from the same sequence. The model takes two different frames, from the same action sequence, separated by $\Delta t$ time to learn the temporal context to the current frame.

Tian~\etal~\cite{tian2013spatiotemporal} uses deformable part based model \cite{felzenszwalb2008discriminatively} to detect the activity in the action frame. Peng~\etal~\cite{peng2016multi} developed a motion region proposal network which was based on faster RCNN \cite{ren2015faster}. two streams (images and optical flow) were used in faster RCNN to generate the activity proposals. Jain~\etal~\cite{jain2014action} use super-voxels to generate activity bounding boxes. The paper produce 2D+t bounding boxes with selective search sampling from the videos. Kalogeiton~\cite{kalogeiton2017action} and \cite{hou2017tube} develop action tube based methods. Both of the models predict the action tube which provides spatial bounding boxes in each of the frame from start to end of an action in a video. Li~\etal~\cite{li2018recurrent} proposed Recurrent Tubelet Proposal and Recognition (RTPR) networks to predict action tubes from start to end of the action in video. The model has two networks, one for proposal and other for recognition. Combination of Convolutional Neural Network (CNN) and Long Short Term Memory (LSTM) Network to learn the recurrent nature of the action proposals.

\section{Problem statement} \label{sec:problem}
\begin{figure}[htb]
    \centering
    \subfloat[\label{fig:p1}]{\includegraphics[width=0.45\textwidth]{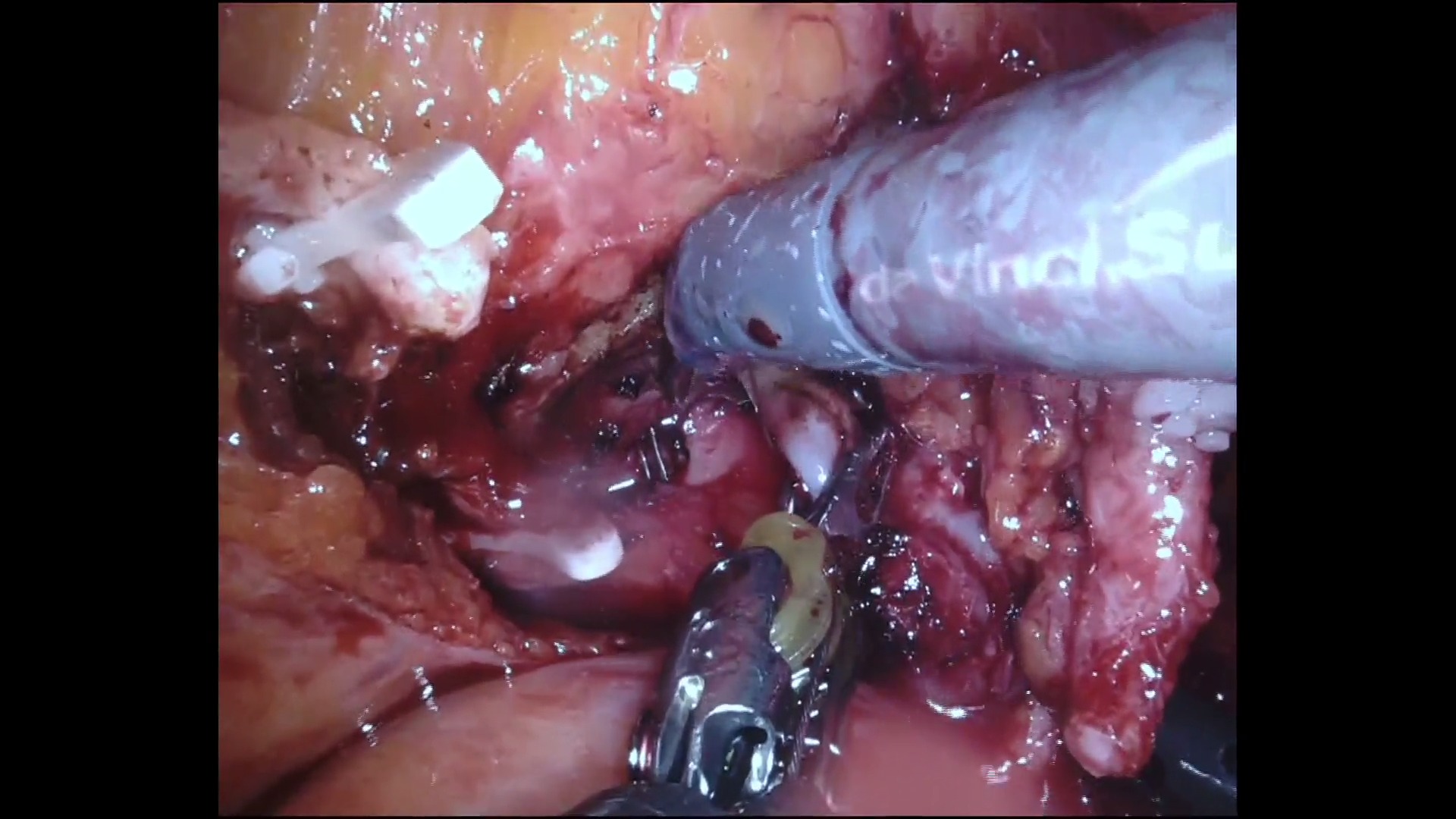}} \hspace{1mm}
    \subfloat[\label{fig:p2}]{\includegraphics[width=0.45\textwidth]{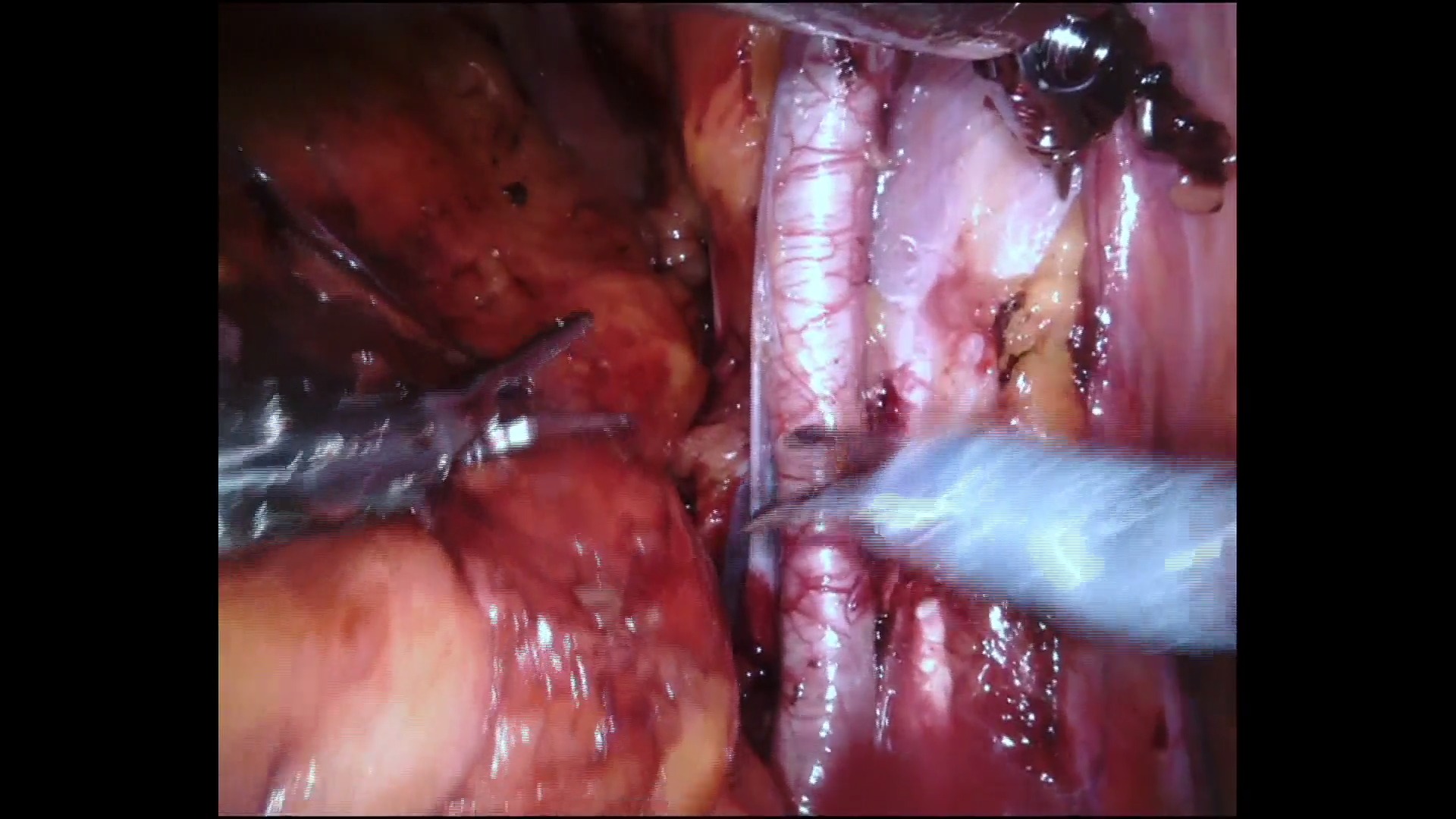}}
    
    \subfloat[\label{fig:p3}]{\includegraphics[width=0.35\textwidth]{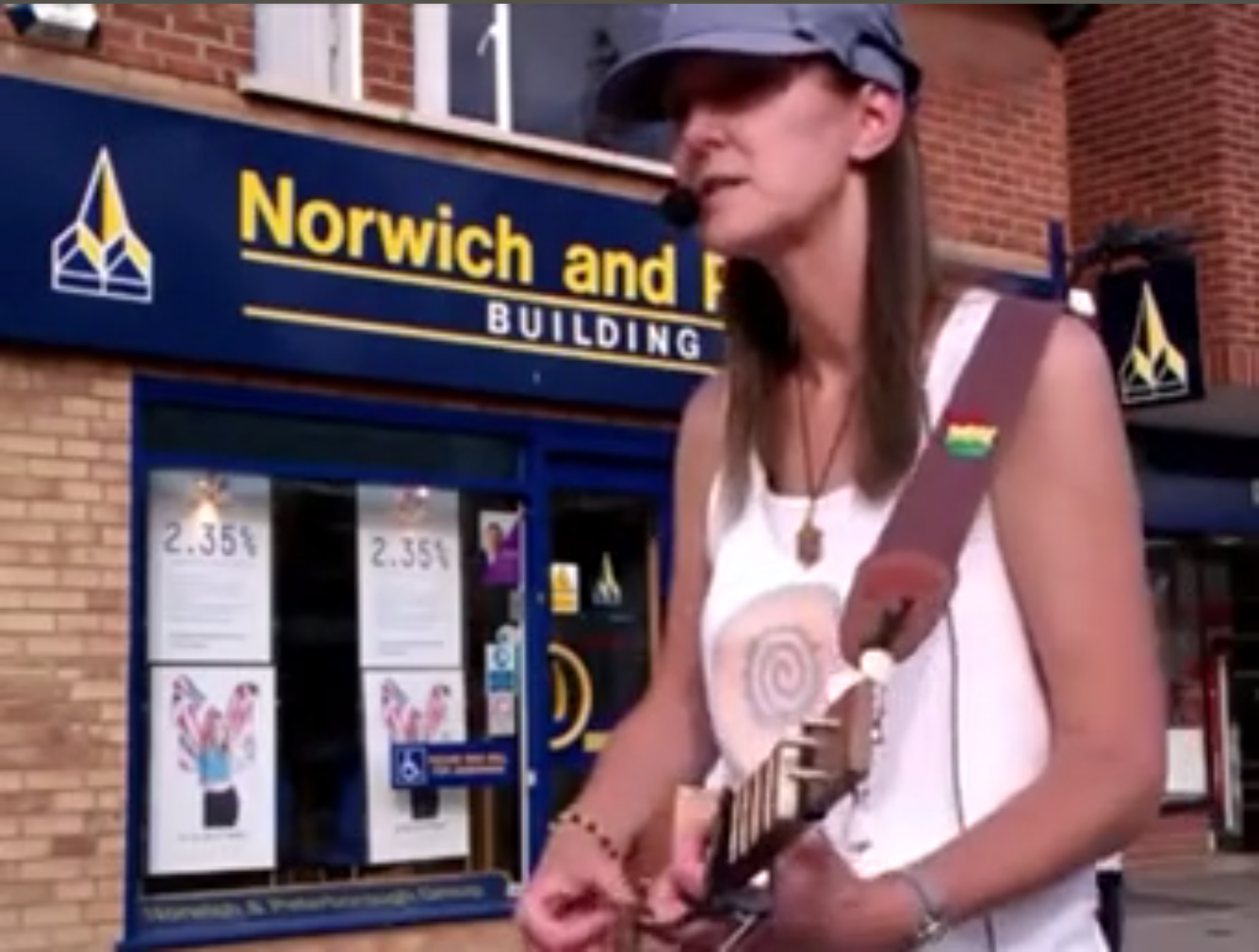}} \hspace{1mm}
    \subfloat[\label{fig:p4}]{\includegraphics[width=0.45\textwidth]{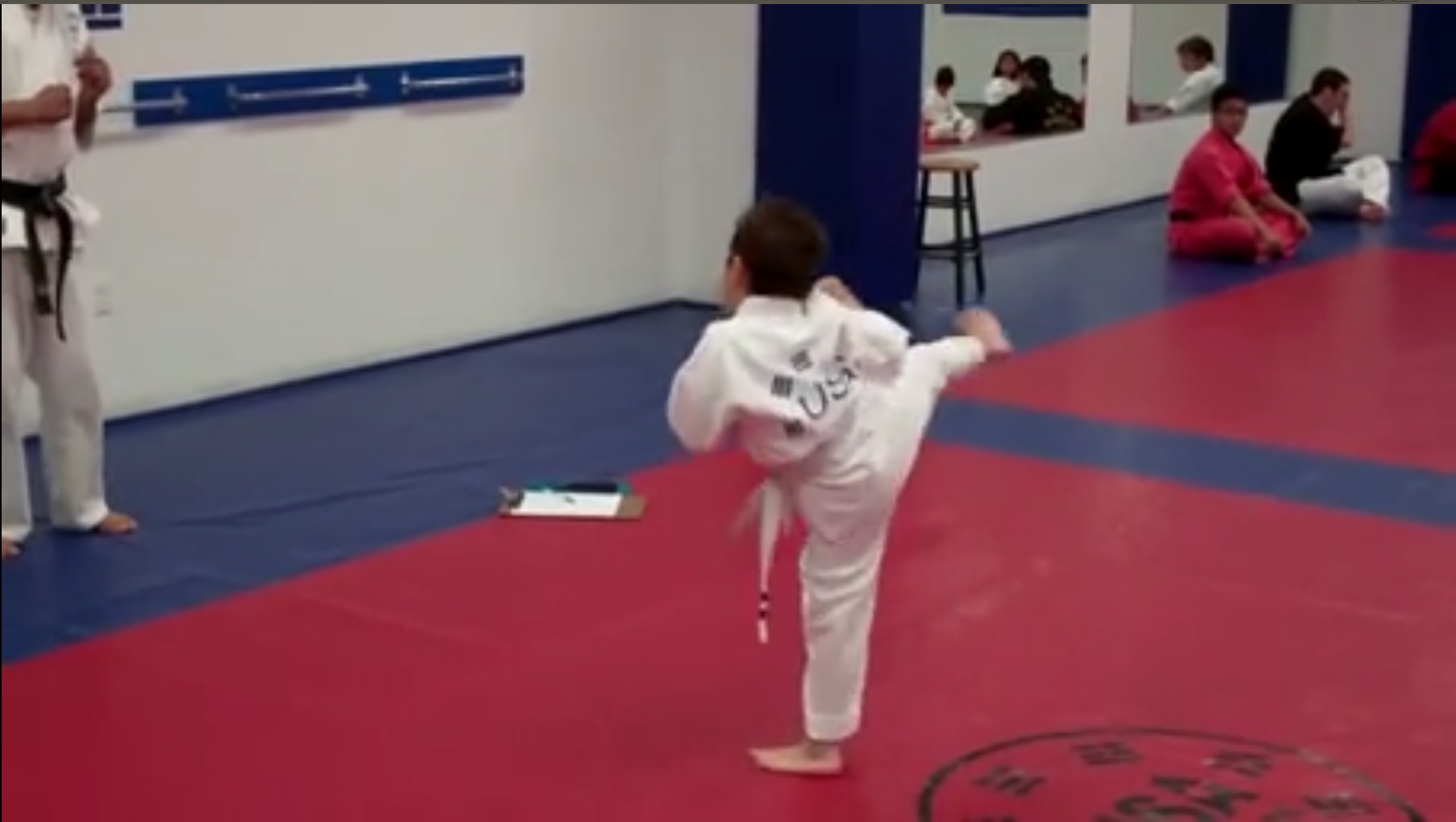}}
    \caption{Image \ref{fig:p1} and \ref{fig:p2} are samples from ESAD dataset. Image \ref{fig:p3} and \ref{fig:p4} are samples from Kinetic-400 dataset~\cite{kay2017kinetics}, which is built from video on YouTube. The difference between the ESAD and Kinetic~\cite{kay2017kinetics} dataset are evident. Endoscope video in ESAD dataset captures images from very close distance loosing all contextual information unlike human activity videos at YouTube. Additionally, general activity dataset like~Kinetic~\cite{kay2017kinetics} or AVA~\cite{gu2018ava} have color, texture, shape and context information making it easier to learn the scene features.}
    \label{fig:problem}
\end{figure}

The task of surgeon action detection is very novel and complex. The key factor that makes this task different from other activity detection task is the appearance of the surgical scene. The most dominant issues with medical images are:

\begin{itemize}
    \item Most important contributor to the difficulty is deformable nature of the organs. As shown in figure \ref{fig:problem}, the organs do not hold a fixed shape in contrast to human activity detection problem, where body has fixed shape and shows a identifiable position. Additionally, boundaries, shapes and color variance between two different organs is minimal, making it very different from standard computer vision tasks.
    \item The scene captured in using endoscope camera is in very close proximity, hence it is unable to show complete organs or its surroundings. hence there is very little contextual information. General activity dataset like~Kinetic~\cite{kay2017kinetics} or AVA~\cite{gu2018ava} have color, texture, shape and context information making it easier to learn the scene features.
    \item Motion and orientation of endoscope in near proximity makes organs appear very different from different angles.
    \item The set of action defined it this dataset provide very accurate description of surgeon actions (e.g., CuttingMesocolon, PullingProstate etc.) to make prediction more informative and useful. Hence in presented dataset, it becomes highly important to know the organ under operation to accurately detect and predict the action.
\end{itemize}

\section{ESAD Dataset} \label{sec:dataset}

\subsubsection{Annotation protocol}

A set of protocols is developed to guide the annotators in their work. This helped minimise the ambiguity in deciding the size of bounding boxes around each action instance, as well as their locations. All annotators were provided a set of instructions with examples in order to standardise the procedure as much as possible. The following guidelines were enforced for the annotation:

\begin{itemize}
    \item Each bounding box should contain both the organ and tool performing the action under consideration, as each action class is highly dependent on the organ under operation.
    \item To balance the presence of tools and organs or tissue in a bounding box, bounding boxes are restricted to containing 30\%-70\% of either tools or organs.
    \item An action label is only assigned when a tool is close enough to the appropriate organ, as informed by the medical expert. Similarly, an action stops as soon as the tool starts to move away from the organ.
    \item Each video frame can have two actions, whose bounding boxes are allowed to overlap.
\end{itemize}

\subsubsection{Structure of dataset}

After a rigorous analysis of the actions performed by surgeons during a typical prostatectomy procedure, we selected 21 action categories for ESAD dataset. Decision is made keeping in mind that action categories should not be too simple that they do not contribute any useful information. Similar, problem is faced by in previous medical action recognition datasets \cite{azari2019using, petlenkov2008application}. Furthermore, the action classes should not be too complex, making it impossible to model the task. 
We concluded the action class list with the help of multiple surgeons and medical professionals. 
Detailed list of classes is shown in table \ref{tab:dataset} along with number of action instances for each category in the whole dataset. 

For the creation of ESAD dataset, we collected four complete prostatectomy procedure with the consent of the patients and hospital. On an average, each video is 2 hours 20 minutes long. Each video is recorded videos  at 30 FPS but annotation are performed at 1 FPS to maintain sufficient variation in the scene. Each frame can have multiple number of action instances. Each instance is annotated with a bounding box and its action label from classes. Surgeon and medical professionals were involved in making the decisions on the appearance of action classes as well as the area of the bounding boxes. As tools operate in close proximity, dataset have a lot of action instances with overlapping bounding boxes. Each annotation is verified by a medical professional.

\begin{figure}[htb]
\center
\subfloat{\includegraphics[width=.49\textwidth]{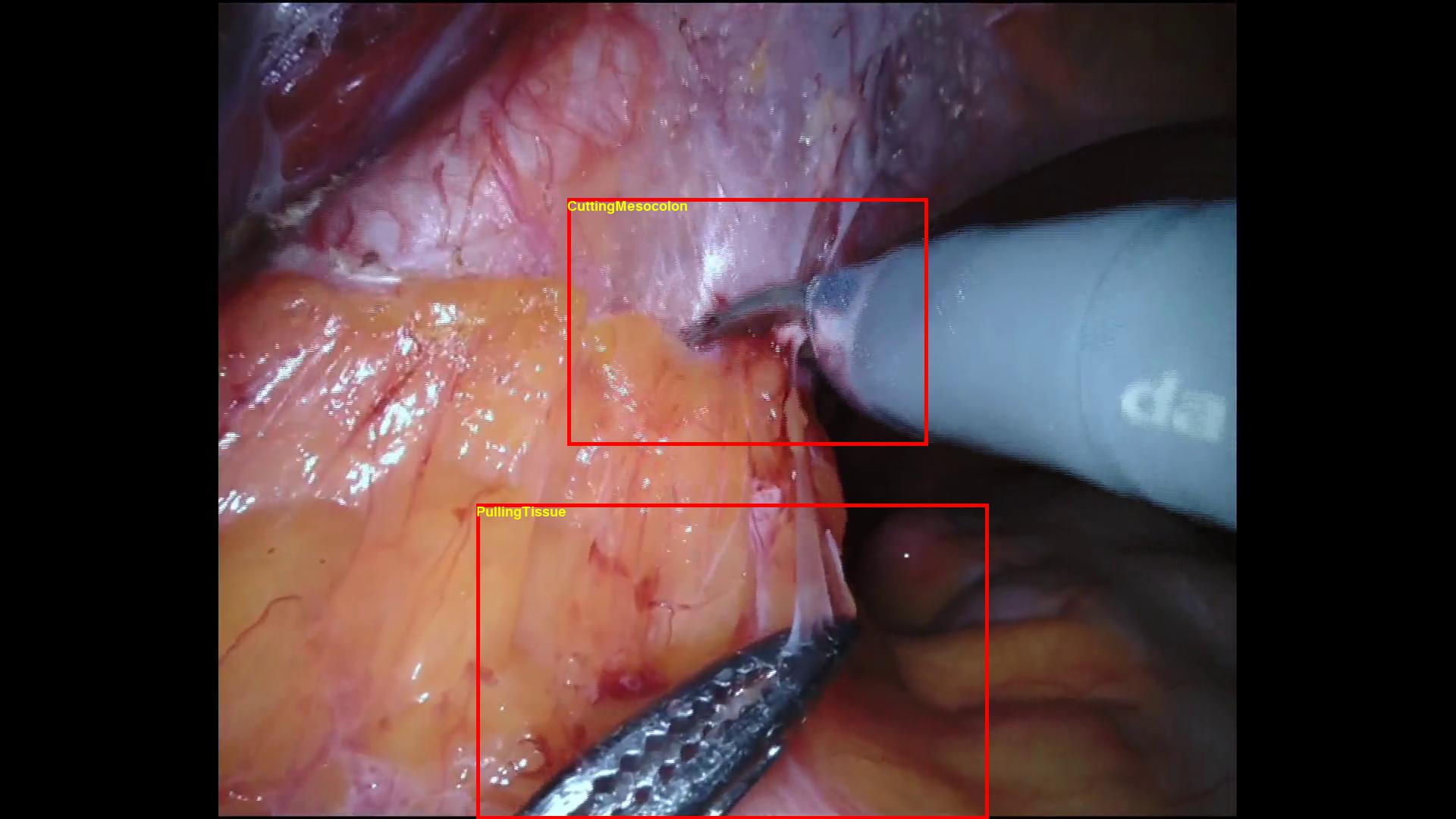}} \hspace{1mm}
\subfloat{\includegraphics[width=.49\textwidth]{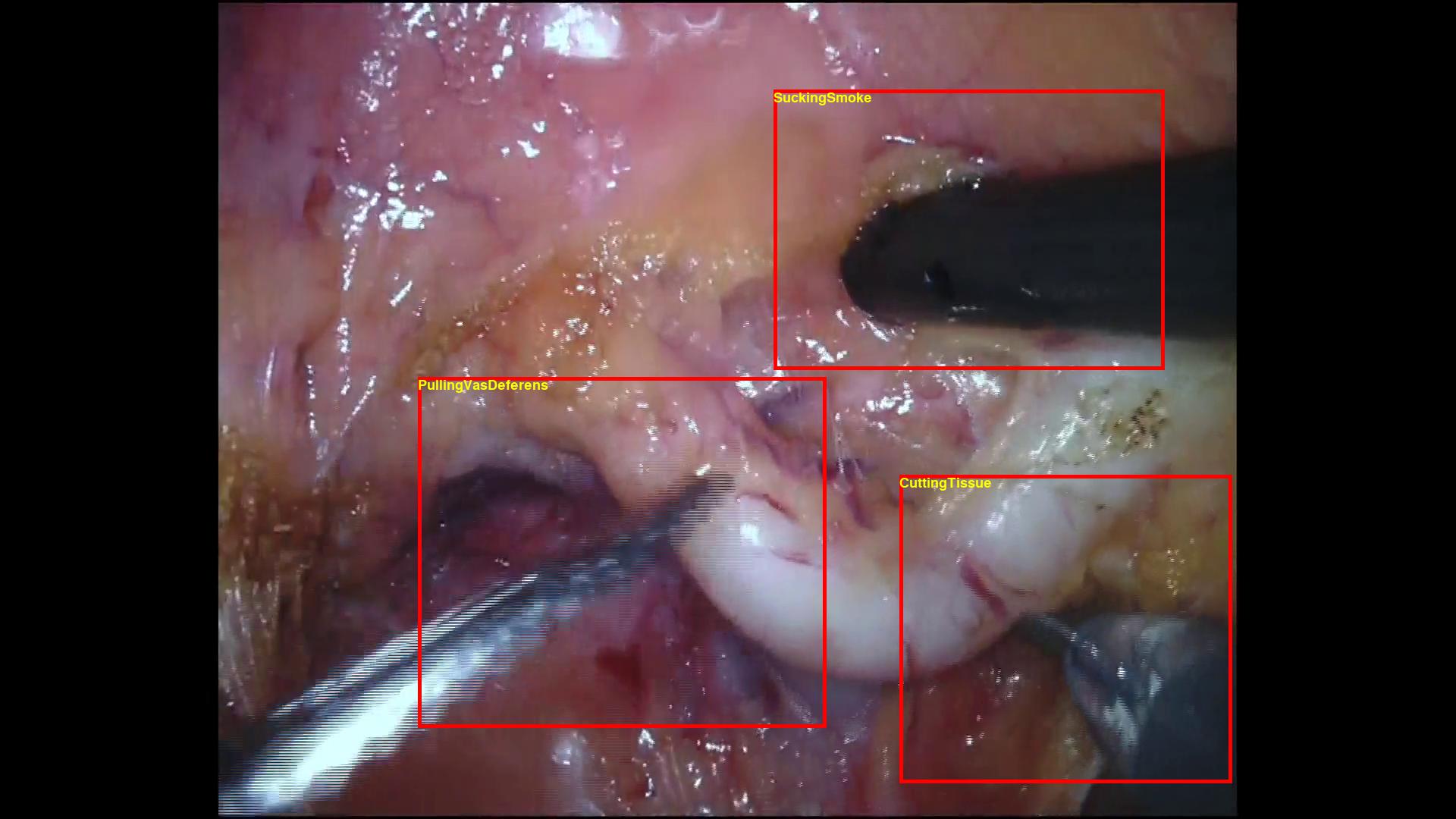}}

\subfloat{\includegraphics[width=.49\textwidth]{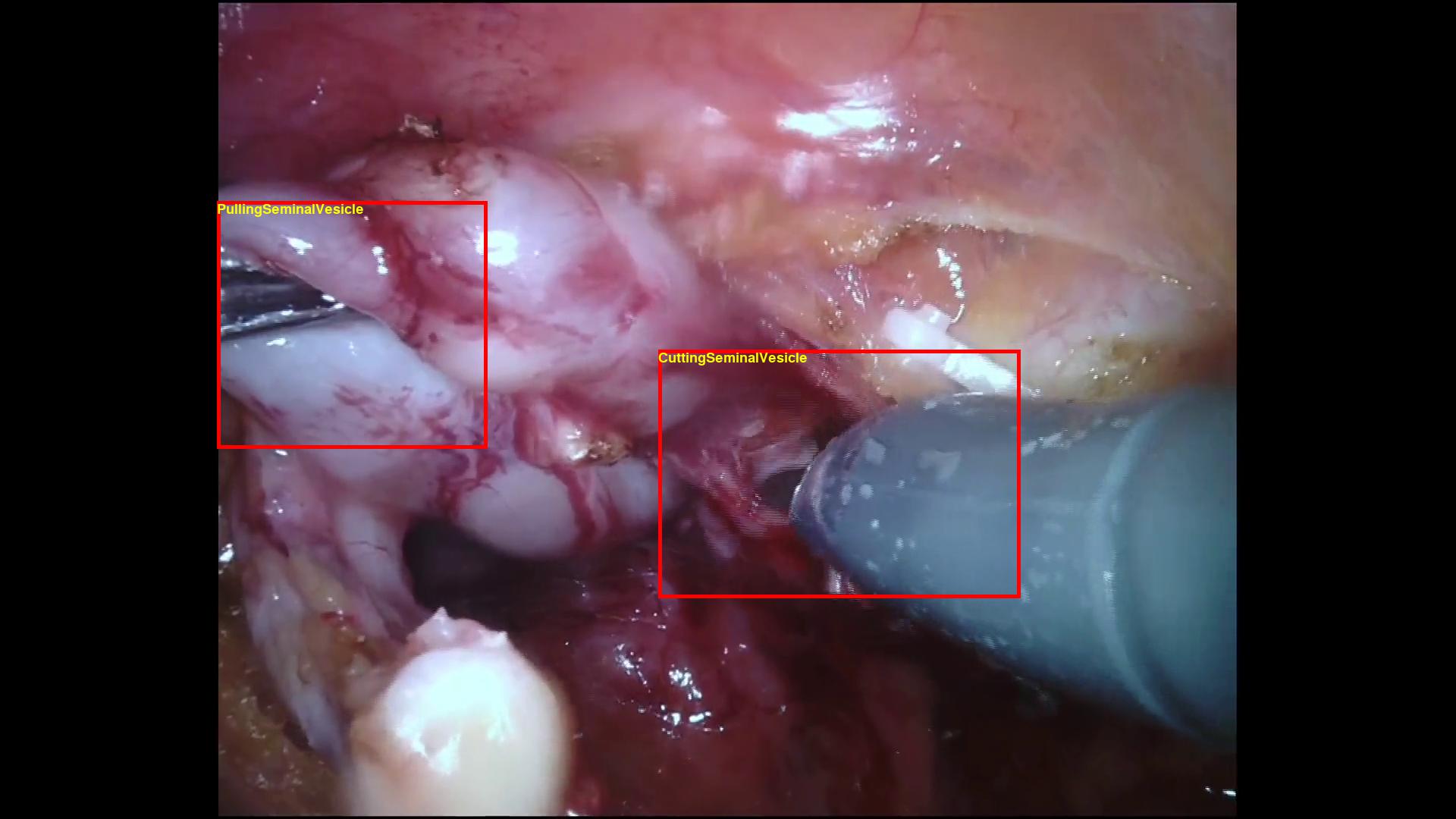}} \hspace{1mm}
\subfloat{\includegraphics[width=.49\textwidth]{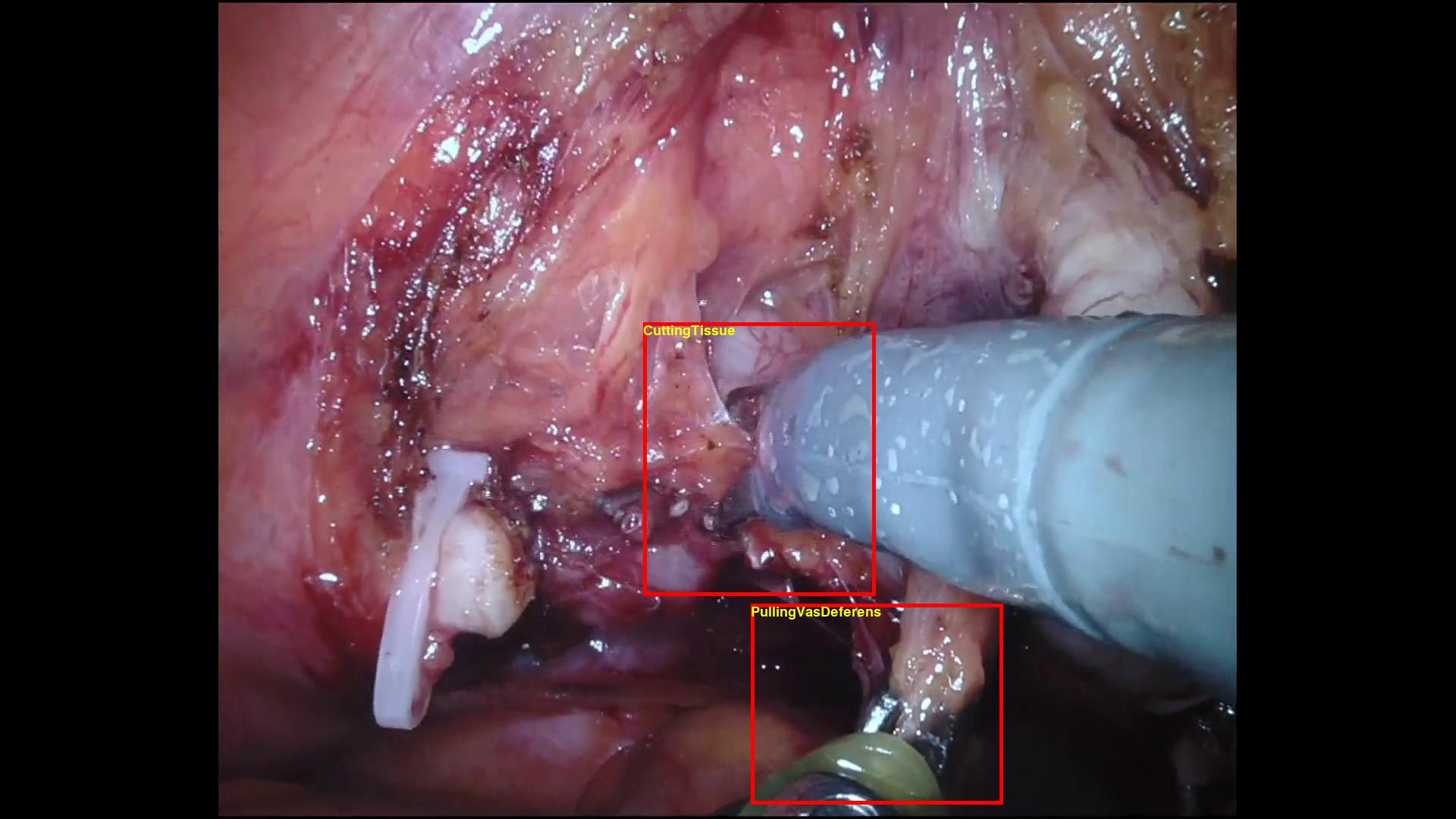}}
\caption{Samples from the ESAD dataset with annotations for surgeon actions. Red box in figure denotes the bounding box. The images are video output from the endoscope during prostatectomy procedure.}
\label{fig:anno_image}
\end{figure}

Some sample from the ESAD dataset are shown in figure \ref{fig:anno_image}. In the images, it can been seen that bounding boxes are centred around the tool as laparoscopic tools represent the subject of action, but dataset also makes sure to include organ under operation. The reasons for that is that most of the surgical action have different names depending on the organ they are operating on despite of the same motion of tools.

\begin{table}[htb]
    \centering
    \begin{tabular}{c|c|c|c|c}
          Label & Train & Val & Test & Total instances \\
         \hline 
         CuttingMesocolon & 315 & 179 & 188 & 682 \\
         PullingVasDeferens & 457 & 245 & 113 & 815 \\
         ClippingVasDeferens & 33 & 25 & 48 & 106 \\
         CuttingVasDeferens & 71 & 22 & 36 & 129 \\
         ClippingTissue & 215 & 44 & 15 & 274 \\
         PullingSeminalVesicle & 2712 & 342 & 436 & 3490 \\
         ClippingSeminalVesicle & 118 & 35 & 33 & 186 \\
         CuttingSeminalVesicle & 2509 & 196 & 307 & 3012 \\
         SuckingBlood & 3753 & 575 & 1696 & 6024 \\
         SuckingSmoke & 381 & 238 & 771 & 1390 \\
         PullingTissue & 4877 & 2177 & 2024 & 9078 \\
         CuttingTissue & 3715 & 1777 & 2055 & 7547 \\
         BaggingProstate & 34 & 5 & 37 & 76 \\
         BladderNeckDissection & 1621 & 283 & 519 & 2423 \\
         BladderAnastomosis & 3585 & 298 & 1828 & 5711 \\
         PullingProstate & 958 & 12 & 451 & 1421 \\
         ClippingBladderNeck & 151 & 24 & 18 & 193 \\
         CuttingThread & 108 & 22 & 40 & 170 \\
         UrethraDissection & 351 & 56 & 439 & 846 \\
         CuttingProstate & 1845 & 56 & 48 & 1949 \\
         PullingBladderNeck & 189 & 509 & 105 & 803 \\
         \hline
    \end{tabular}
    \caption{List of actions for ESAD dataset with number of samples for training, validation and test.}
    \label{tab:dataset}
\end{table}

\begin{figure}[htb]
    \centering
    \includegraphics{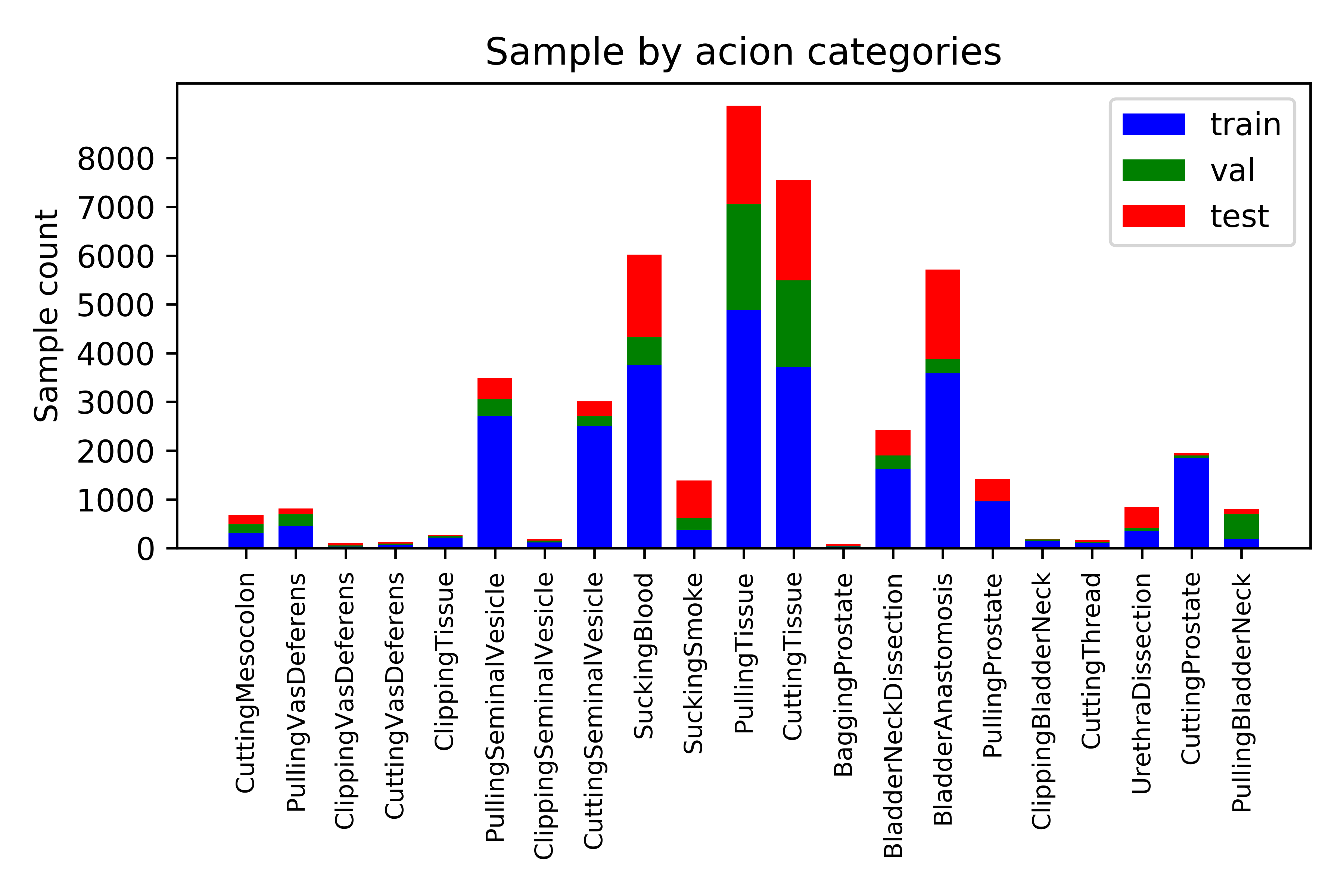}
    \caption{Distribution of samples for all action categories in training, validation and test splits. Blue, green and red colors of the bars represent training, validation and test set of ESAD.}
    \label{fig:esad_dist}
\end{figure}

\subsubsection{Dataset split}

The dataset is divided into three different sets: training, validation and test. The two surgeries with with the maximum number of action instances were selected as training set. The one video with the most balanced number of samples for each action class was used as the test set. The objective of dataset is to provide a fair evaluation to all types of algorithms. Hence, the last procedure is selected for validation set. The number of instances (labelled bounding boxes) for each action class can be seen in Table \ref{tab:dataset}.

Distribution of samples for each action category for each of the three splits is shown in figure \ref{fig:esad_dist}. It is clear from the bar chart the dataset is highly skewed in term of class imbalance. The reason for this is the nature of surgical procedures. As shown in figure \ref{fig:esad_dist}, classes \textit{PullingTissue} and \textit{CuttingTissue} contain higest number of sampels as this is the most common action performed by surgeon during prostatectomy. Whereas, classes like \textit{BaggingProstate} and \textit{CuttingThread} have lowest samples due to short duration of these activities per procedure.

\section{Results and discussion}~\label{sec:results}

In this section, we start by presenting a baseline model~\ref{subsec:baseline_model} used to establish a baseline on our dataset. 
We show results of baseline mode in ~\ref{subsec:results} section with evaluation metric described in~\ref{subsec:eval-metric}. 
Finally, we will discuss the understanding gained from the results~\ref{subsec:results} in ~\ref{subsec:discussion}.

\subsection{Baseline model}~\label{subsec:baseline_model}
The baseline model is based on Feature Pyramidal Network (FPN) architecture. The concept was originally proposed by Lin \etal~\cite{lin2017feature}. The paper uses convolutional CNN architecture with pooling layers. Residual networks (ResNet)~\cite{he2016deep} is used as a backbone network for the detection model. Output of each residual block is used to build the pyramid features. Residual feature maps on different level of pyramid are then feed to a sub-net made of 4 convolutional layers and finally a convolutional layer to predict the class scores and bounding box coordinated respectively. Similar to the original paper~\cite{lin2017feature}, we freeze the batch normalisation layers of ResNet based backbone networks. Also, few initial layers are also frozen to avoid the overfitting. Finally, non maximum suppression (NMS) is used to discard the false positives in the model predictions at the test time.

We try two different loss functions to train classification sub-net of our baseline model. Since our baseline model is single stage model and based on  \cite{lin2017feature}, following~ \cite{lin2017feature}, we train FPN with online hard example mining (OHEM)-loss \cite{liu2016ssd} and Focal loss \cite{lin2017focal}. We use smooth-l1 loss~\cite{ren2015faster} to train regression sub-net.

\paragraph{Implementation details:} We train baseline model with various input image size, e.g. 200, 400 or 600. Short size of the input image is resized to input image size and longer side is resized with same scale. We set the learning rate $0.01$ and batch size to $16$. The networks are trained for $7K$ iteration with learning rate drop by the factor of $10$ after $5K$ iterations. The complete model is implemented in \href{https://pytorch.org/}{pytorch} and is provided as open access at \url{https://github.com/Viveksbawa/SARAS-ESAD-Baseline}. At the moment, the source code supports pytorch1.5 and Ubuntu with Anaconda distribution of python. It is tested on machines with 2/4/8 GPUs.

\subsection{Evaluation metric}~\label{subsec:eval-metric}
We also used three different IOU thresholds to compute the average precision (AP). 
We AP computed at 0.1, 0.3 and 0.5 AP are named as $AP_{10}$, $AP_{30}$ and $AP_{50}$, respectively. Then, mean is computed at three thresholds to get a final evaluation score. the purpose of computing three different APs is to capture quality of detection as well as classification. As we know this is a new and complex task, it is very difficult to get good detection accuracy at higher threshold (can be seen in $AP_{50}$ column). Hence, we want to identify both- how accurately model can detect the classes of actions present in the scene as well as the location of their bounding boxes. 

\subsection{Results}~\label{subsec:results}
The results achieved by model with both of the losses are shown in table \ref{tab:esad_results_ResNet50}. 
We trained model on four different image sizes: 200, 400, 600, 800. 
Motive behind it is to observe the effect of tool sizes in the image on the detection accuracy.
As we can observe in the table, with increase in image size, models with OHEM loss functions is able to achieve better detection accuracy on validation set. While the same can not said for test-set. 

\begin{table}[]
    \centering
    \begin{tabular}{lcccccc}
    \toprule
        Loss & image size & $AP_{10}$ & $AP_{30}$ & $AP_{50}$ & $AP_{mean}$ & $Test-AP_{mean}$ \\ \midrule
        Focal & 200 & 33.8 & 17.7 & 6.6 & 19.4 & 15.7 \\
        Focal & 400 & 35.9 & 19.4 & 8.0 & 21.1 & \textbf{16.1} \\
        Focal & 600 & 29.2 & 17.6 & 8.7 & 18.5 & 14.0 \\
        Focal & 800 & 31.9 & 20.1 & 8.7 & 20.2 & 12.4 \\
        OHEM & 200 & 35.1 & 18.7 & 6.3 & 20.0 & 11.3 \\
        OHEM & 400 & 33.9 & 19.2 & 7.4 & 20.2 & 13.6 \\
        OHEM & 600 & 37.6 & 23.4 & 11.2 & 24.1 & 12.5 \\
        OHEM & 800 & \textbf{36.8} & \textbf{24.3} & \textbf{12.2} & \textbf{24.4} & 12.3 \\
        \bottomrule
    \end{tabular}
    \caption{Results of the baseline models with different loss function and input image sizes, where  backbone network fixed to ResNet50. $AP_{10}$, $AP_{30}$, $AP_{50}$, and $AP_{mean}$ are presented on validation-set, while $Test-AP_{mean}$ is computed based on test-set similar to $AP_{mean}$.}
    \label{tab:esad_results_ResNet50}
\end{table}

\begin{table}[]
    \centering
    \begin{tabular}{lcccccc}
    \toprule
    Loss & backbone & $AP_{10}$ & $AP_{30}$ & $AP_{50}$ & $AP_{mean}$ & $Test-AP_{mean}$ \\ \midrule
    Focal & ResNet18 & 35.1 & 18.9 & 8.1 & 20.7 & \textbf{15.3} \\
    OHEM & ResNet18 & 36.0 & \textbf{20.7} & 7.7 & \textbf{21.5} & 13.8 \\
    Focal & ResNet34 & 34.6 & 18.9 & 6.4 & 19.9 & 14.3 \\
    OHEM & ResNet34 & \textbf{36.7} & 20.4 & 7.1 & 21.4 & 13.8 \\
    Focal & ResNet50 & 35.9 & 19.4 & \textbf{8.0} & 21.1 & 16.1 \\
    OHEM & ResNet50 & 33.9 & 19.2 & 7.4 & 20.2 & 13.6 \\
    Focal & ResNet101 & 32.5 & 17.2 & 6.1 & 18.6 & 14.0 \\
    OHEM & ResNet101 & 36.6 & 20.1 & 7.4 & 21.3 & 12.3 \\
    \bottomrule
    \end{tabular}
    \caption{Results of the baseline models with different loss function, backbone networks, where input image size is fixed to 400. $AP_{10}$, $AP_{30}$, $AP_{50}$, and $AP_{mean}$ are presented on validation-set, while $Test-AP_{mean}$ is computed based on test-set similar to $AP_{mean}$.}
    \label{tab:esad_results_ims400}
\end{table}

The table \ref{tab:esad_results_ims400} shows the results achieved by base model with different backbone networks while keeping the input image size fixed to 400 on both validation and test sets. It is clear from Tables~\ref{tab:esad_results_ims400} and~\ref{tab:esad_results_ResNet50} that OHEM loss performs better at validation set and focal loss performs better at test set.

From above results, it is clear that the presented baseline method is still far from achieve satisfactory performance. 
We hope that this will server as good benchmark for future methods which are specifically designed for endoscopic video. 


\section{Conclusion}
This paper presents first of its kind dataset for action detection in surgical images.
The dataset is developed on real videos collected from the prostatectomy procedures. 
This dataset aim to provide a benchmark for medical computer vision community to develop and test the state of the art algorithms for surgical robotics. 
We also released a baseline model along with the dataset which is developed using fully convolutional network architecture. Model is tested with two different type of loss functions: online hard example mining and focal loss. Focal loss based model is able to generalize much better for the test set. Additionally, we found out that bigger images size results in better model performance, generally. Complexity of dataset is also evaluated with different backbone architectures. Medium complexity/depth models like ResNet-34 perform better that higher depth models.  

\section{Acknowledgement}
This work is conducted under the European Union’s Horizon 2020 research and innovation programme under grant agreement No. 779813 and name of the project is Smart Autonomous Robotic Assistant Surgeon (SARAS project).

\bibliographystyle{plainnat}
\bibliography{reference}

\begin{thebibliography}{26}
\providecommand{\natexlab}[1]{#1}
\providecommand{\url}[1]{\texttt{#1}}
\expandafter\ifx\csname urlstyle\endcsname\relax
  \providecommand{\doi}[1]{doi: #1}\else
  \providecommand{\doi}{doi: \begingroup \urlstyle{rm}\Url}\fi

\bibitem[Azari et~al.(2019)Azari, Hu, Miller, Le, and Radwin]{azari2019using}
David~P Azari, Yu~Hen Hu, Brady~L Miller, Brian~V Le, and Robert~G Radwin.
\newblock Using surgeon hand motions to predict surgical maneuvers.
\newblock \emph{Human factors}, 61\penalty0 (8):\penalty0 1326--1339, 2019.

\bibitem[Felzenszwalb et~al.(2008)Felzenszwalb, McAllester, and
  Ramanan]{felzenszwalb2008discriminatively}
Pedro Felzenszwalb, David McAllester, and Deva Ramanan.
\newblock A discriminatively trained, multiscale, deformable part model.
\newblock In \emph{2008 IEEE Conference on Computer Vision and Pattern
  Recognition}, pages 1--8. IEEE, 2008.

\bibitem[Gkioxari et~al.(2015)Gkioxari, Girshick, and
  Malik]{gkioxari2015contextual}
Georgia Gkioxari, Ross Girshick, and Jitendra Malik.
\newblock Contextual action recognition with r* cnn.
\newblock In \emph{Proceedings of the IEEE international conference on computer
  vision}, pages 1080--1088, 2015.

\bibitem[Gu et~al.(2018)Gu, Sun, Ross, Vondrick, Pantofaru, Li,
  Vijayanarasimhan, Toderici, Ricco, Sukthankar, et~al.]{gu2018ava}
Chunhui Gu, Chen Sun, David~A Ross, Carl Vondrick, Caroline Pantofaru, Yeqing
  Li, Sudheendra Vijayanarasimhan, George Toderici, Susanna Ricco, Rahul
  Sukthankar, et~al.
\newblock Ava: A video dataset of spatio-temporally localized atomic visual
  actions.
\newblock In \emph{Proceedings of the IEEE Conference on Computer Vision and
  Pattern Recognition}, pages 6047--6056, 2018.

\bibitem[He et~al.(2016)He, Zhang, Ren, and Sun]{he2016deep}
Kaiming He, Xiangyu Zhang, Shaoqing Ren, and Jian Sun.
\newblock Deep residual learning for image recognition.
\newblock In \emph{Proceedings of the IEEE conference on computer vision and
  pattern recognition}, pages 770--778, 2016.

\bibitem[Hou et~al.(2017{\natexlab{a}})Hou, Chen, and Shah]{hou2017end}
Rui Hou, Chen Chen, and Mubarak Shah.
\newblock An end-to-end 3d convolutional neural network for action detection
  and segmentation in videos.
\newblock \emph{arXiv preprint arXiv:1712.01111}, 2017{\natexlab{a}}.

\bibitem[Hou et~al.(2017{\natexlab{b}})Hou, Chen, and Shah]{hou2017tube}
Rui Hou, Chen Chen, and Mubarak Shah.
\newblock Tube convolutional neural network (t-cnn) for action detection in
  videos.
\newblock In \emph{Proceedings of the IEEE International Conference on Computer
  Vision}, pages 5822--5831, 2017{\natexlab{b}}.

\bibitem[Jain et~al.(2014)Jain, Van~Gemert, J{\'e}gou, Bouthemy, and
  Snoek]{jain2014action}
Mihir Jain, Jan Van~Gemert, Herv{\'e} J{\'e}gou, Patrick Bouthemy, and Cees~GM
  Snoek.
\newblock Action localization with tubelets from motion.
\newblock In \emph{Proceedings of the IEEE conference on computer vision and
  pattern recognition}, pages 740--747, 2014.

\bibitem[Kalogeiton et~al.(2017)Kalogeiton, Weinzaepfel, Ferrari, and
  Schmid]{kalogeiton2017action}
Vicky Kalogeiton, Philippe Weinzaepfel, Vittorio Ferrari, and Cordelia Schmid.
\newblock Action tubelet detector for spatio-temporal action localization.
\newblock In \emph{Proceedings of the IEEE International Conference on Computer
  Vision}, pages 4405--4413, 2017.

\bibitem[Kay et~al.(2017)Kay, Carreira, Simonyan, Zhang, Hillier,
  Vijayanarasimhan, Viola, Green, Back, Natsev, et~al.]{kay2017kinetics}
Will Kay, Joao Carreira, Karen Simonyan, Brian Zhang, Chloe Hillier, Sudheendra
  Vijayanarasimhan, Fabio Viola, Tim Green, Trevor Back, Paul Natsev, et~al.
\newblock The kinetics human action video dataset.
\newblock \emph{arXiv preprint arXiv:1705.06950}, 2017.

\bibitem[Kocev et~al.(2014)Kocev, Ritter, and Linsen]{kocev2014projector}
Bojan Kocev, Felix Ritter, and Lars Linsen.
\newblock Projector-based surgeon--computer interaction on deformable surfaces.
\newblock \emph{International journal of computer assisted radiology and
  surgery}, 9\penalty0 (2):\penalty0 301--312, 2014.

\bibitem[Li et~al.(2018)Li, Qiu, Dai, Yao, and Mei]{li2018recurrent}
Dong Li, Zhaofan Qiu, Qi~Dai, Ting Yao, and Tao Mei.
\newblock Recurrent tubelet proposal and recognition networks for action
  detection.
\newblock In \emph{Proceedings of the European conference on computer vision
  (ECCV)}, pages 303--318, 2018.

\bibitem[LI et~al.(2016)LI, OHYA, CHIBA, XU, and YAMASHITA]{li2016subaction}
Ye~LI, Jun OHYA, Toshio CHIBA, Rong XU, and Hiromasa YAMASHITA.
\newblock Subaction based early recognition of surgeons' hand actions from
  continuous surgery videos.
\newblock \emph{IIEEJ transactions on image electronics and visual computing},
  4\penalty0 (2):\penalty0 124--135, 2016.

\bibitem[Lin et~al.(2017{\natexlab{a}})Lin, Doll{\'a}r, Girshick, He,
  Hariharan, and Belongie]{lin2017feature}
Tsung-Yi Lin, Piotr Doll{\'a}r, Ross Girshick, Kaiming He, Bharath Hariharan,
  and Serge Belongie.
\newblock Feature pyramid networks for object detection.
\newblock In \emph{Proceedings of the IEEE conference on computer vision and
  pattern recognition}, pages 2117--2125, 2017{\natexlab{a}}.

\bibitem[Lin et~al.(2017{\natexlab{b}})Lin, Goyal, Girshick, He, and
  Doll{\'a}r]{lin2017focal}
Tsung-Yi Lin, Priya Goyal, Ross Girshick, Kaiming He, and Piotr Doll{\'a}r.
\newblock Focal loss for dense object detection.
\newblock In \emph{Proceedings of the IEEE international conference on computer
  vision}, pages 2980--2988, 2017{\natexlab{b}}.

\bibitem[Liu et~al.(2016)Liu, Anguelov, Erhan, Szegedy, Reed, Fu, and
  Berg]{liu2016ssd}
Wei Liu, Dragomir Anguelov, Dumitru Erhan, Christian Szegedy, Scott Reed,
  Cheng-Yang Fu, and Alexander~C Berg.
\newblock Ssd: Single shot multibox detector.
\newblock In \emph{European conference on computer vision}, pages 21--37.
  Springer, 2016.

\bibitem[Makary and Daniel(2016)]{jhustudy}
Martin Makary and Michael Daniel.
\newblock Study suggests medical errors now third leading cause of death in the
  u.s.
\newblock
  \url{https://www.hopkinsmedicine.org/news/media/releases/study_suggests_medical_errors_now_third_leading_cause_of_death_in_the_us},
  2016.
\newblock Online; accessed 15-April-2020.

\bibitem[Nepogodiev et~al.(2019)Nepogodiev, Martin, Biccard, Makupe, Bhangu,
  Ademuyiwa, et~al.]{nepogodiev2019global}
Dmitri Nepogodiev, Janet Martin, Bruce Biccard, Alex Makupe, Aneel Bhangu,
  Adesoji Ademuyiwa, et~al.
\newblock Global burden of postoperative death.
\newblock \emph{Lancet}, 393\penalty0 (401):\penalty0 33139--8, 2019.

\bibitem[Peng and Schmid(2016)]{peng2016multi}
Xiaojiang Peng and Cordelia Schmid.
\newblock Multi-region two-stream r-cnn for action detection.
\newblock In \emph{European conference on computer vision}, pages 744--759.
  Springer, 2016.

\bibitem[Petlenkov et~al.(2008)Petlenkov, Nomm, Vain, and
  Miyawaki]{petlenkov2008application}
Eduard Petlenkov, Sven Nomm, Juri Vain, and Fujio Miyawaki.
\newblock Application of self organizing kohonen map to detection of surgeon
  motions during endoscopic surgery.
\newblock In \emph{2008 IEEE International Joint Conference on Neural Networks
  (IEEE World Congress on Computational Intelligence)}, pages 2806--2811. IEEE,
  2008.

\bibitem[Ren et~al.(2015)Ren, He, Girshick, and Sun]{ren2015faster}
Shaoqing Ren, Kaiming He, Ross Girshick, and Jian Sun.
\newblock Faster r-cnn: Towards real-time object detection with region proposal
  networks.
\newblock In \emph{Advances in neural information processing systems}, pages
  91--99, 2015.

\bibitem[Saha et~al.(2017)Saha, Singh, and Cuzzolin]{saha2017amtnet}
Suman Saha, Gurkirt Singh, and Fabio Cuzzolin.
\newblock Amtnet: Action-micro-tube regression by end-to-end trainable deep
  architecture.
\newblock In \emph{Proceedings of the IEEE International Conference on Computer
  Vision}, pages 4414--4423, 2017.

\bibitem[Singh et~al.(2017)Singh, Saha, Sapienza, Torr, and
  Cuzzolin]{singh2017online}
Gurkirt Singh, Suman Saha, Michael Sapienza, Philip~HS Torr, and Fabio
  Cuzzolin.
\newblock Online real-time multiple spatiotemporal action localisation and
  prediction.
\newblock In \emph{Proceedings of the IEEE International Conference on Computer
  Vision}, pages 3637--3646, 2017.

\bibitem[Tian et~al.(2013)Tian, Sukthankar, and Shah]{tian2013spatiotemporal}
Yicong Tian, Rahul Sukthankar, and Mubarak Shah.
\newblock Spatiotemporal deformable part models for action detection.
\newblock In \emph{Proceedings of the IEEE conference on computer vision and
  pattern recognition}, pages 2642--2649, 2013.

\bibitem[van Amsterdam et~al.(2019)van Amsterdam, Nakawala, De~Momi, and
  Stoyanov]{van2019weakly}
Beatrice van Amsterdam, Hirenkumar Nakawala, Elena De~Momi, and Danail
  Stoyanov.
\newblock Weakly supervised recognition of surgical gestures.
\newblock In \emph{2019 International Conference on Robotics and Automation
  (ICRA)}, pages 9565--9571. IEEE, 2019.

\bibitem[Voros and Hager(2008)]{voros2008towards}
Sandrine Voros and Gregory~D Hager.
\newblock Towards “real-time” tool-tissue interaction detection in
  robotically assisted laparoscopy.
\newblock In \emph{2008 2nd IEEE RAS \& EMBS International Conference on
  Biomedical Robotics and Biomechatronics}, pages 562--567. IEEE, 2008.

\end{thebibliography}


\end{document}